\documentclass{article} 
\PassOptionsToPackage{dvipsnames}{xcolor}
\usepackage{iclr2025_conference,times}

\usepackage{subcaption}
\usepackage{booktabs}
\usepackage{multirow}
\usepackage{tabularx}
\usepackage{float}
\usepackage{graphicx}
\usepackage{caption}
\usepackage[margin=1in]{geometry}

\usepackage{listings}
\definecolor{codegreen}{rgb}{0,0.6,0}
\definecolor{codegray}{rgb}{0.5,0.5,0.5}
\definecolor{codepurple}{rgb}{0.58,0,0.82}
\definecolor{backcolour}{rgb}{0.95,0.95,0.92}
\lstdefinestyle{mystyle}{
    backgroundcolor=\color{backcolour},
    commentstyle=\color{codegreen},
    keywordstyle=\color{magenta},
    numberstyle=\tiny\color{codegray},
    stringstyle=\color{codepurple},
    basicstyle=\ttfamily\footnotesize,
    breaklines=true,
    breakindent=0pt,
    keepspaces=true,
    numbersep=5pt,
    showspaces=false,
    showstringspaces=false,
    frame=single,
    columns=fullflexible,
    literate=
    {•}{{*}}1
    {–}{{-}}1
    {—}{{-}}1
    {‐}{{-}}1
    {-}{{-}}1
    {“}{{"}}1
    {”}{{"}}1
    {‘}{{'}}1
    {’}{{'}}1
    {…}{{...}}1
    {⇒}{{$\Rightarrow$}}1
    {≤}{{$\leq$}}1
    {→}{{$\rightarrow$}}1
}
\usepackage{amsmath,amssymb,amsthm,mathtools,bm,bbm}
\usepackage{enumitem}
\usepackage{hyperref}
\usepackage{array}
\newcolumntype{C}[1]{>{\centering\arraybackslash}m{#1}}
\usepackage{pgfplots}
\pgfplotsset{compat=1.18}
\usepackage{wrapfig}

\newcommand{\1}{\mathbf{1}}

\definecolor{mybarcolor}{RGB}{77,171,247}

\usepackage{tikz}
\usetikzlibrary{arrows.meta,positioning,calc,shapes.geometric}

\usepackage[most]{tcolorbox}
\tcbuselibrary{listingsutf8}
\definecolor{promptbg}{RGB}{233,240,255}
\definecolor{annotbg}{RGB}{255,250,200}
\definecolor{highlight}{RGB}{252,213,122}

\newtcolorbox{instructbox}[1][]{
  breakable,
  enhanced,
  colback=gray!4,
  colframe=black!50,
  sharp corners,
  fontupper=\small\ttfamily,
  left=8pt,
  right=8pt,
  top=6pt,
  bottom=6pt,
  boxrule=0.6pt,
  listing only,
  listing options={
    basicstyle=\ttfamily\small,
    breaklines=true,
    postbreak=\mbox{\textcolor{gray}{$\hookrightarrow$}\space},
    showstringspaces=false,
    tabsize=2,
    numbers=none,
    stepnumber=1,
    xleftmargin=0pt,
    xrightmargin=0pt,
    frame=none,
    keywordstyle=\color{blue}\bfseries,
    emph={curr_instructions,inputs_outputs_feedback},
    emphstyle=\color{teal}
  },
  title=#1
}

\usepackage{arydshln}
\usepackage{algorithm}
\usepackage{algpseudocode}

\usepackage{amsmath,amsfonts,bm}

\def\eqref#1{equation~\ref{#1}}

\def\1{\bm{1}}

\DeclareMathAlphabet{\mathsfit}{\encodingdefault}{\sfdefault}{m}{sl}
\SetMathAlphabet{\mathsfit}{bold}{\encodingdefault}{\sfdefault}{bx}{n}

\usepackage{url}

\title{Do Agent Optimizers Compound?\\
A Continual-Learning Evaluation on Terminal-Bench 2.0\thanks{Baseline, Phase 1, and Phase 2 agent artifacts for all three optimizers are released at \url{https://github.com/relai-ai/Continual-Learning-Terminal-Bench}.}}

\author{
Wenxiao Wang \\
\texttt{wwx@relai.ai}\\
RELAI.ai\\
\And Priyatham Kattakinda \\
\texttt{priyatham@relai.ai}\\
RELAI.ai\\
\And Soheil Feizi\\
\texttt{sfeizi@relai.ai}\\
RELAI.ai\\
}

\date{}

\iclrfinalcopy

\begin{document}

\maketitle

\begin{abstract}
Most reported gains from agent-optimization methods are one-shot: an agent is optimized against a fixed benchmark and the resulting improvement is reported as if it were a stable property of the method. This does not test the setting that matters for deployed agents, where optimization is applied recursively as new failures and new tasks appear over time. The central question this raises is whether optimizer-driven gains \emph{compound}: after an agent has been optimized once, can it be optimized again on newly arrived tasks without eroding the gains the first round produced? We study this question with a two-phase continual-learning evaluation built from hard tasks in Terminal-Bench 2.0, comparing three approaches to agent-harness optimization (GEPA, Meta Harness, and RELAI's Verifiable Continual Learning, RELAI-VCL) under identical optimization budgets. All three methods improve over the baseline agent in the conventional, static, single-phase setting. However, once new tasks are introduced, the methods diverge sharply: GEPA's optimized agent transfers below the unoptimized baseline, Meta Harness transfers well but fails to improve further once given a second optimization budget, and RELAI-VCL is the only method that both transfers positively to unseen tasks and continues improving after those tasks are folded into the optimization objective, reaching the highest pass rate at every evaluated stage and the highest lifelong average pass rate overall (76.4\% vs.\ 66.0\% for GEPA, 64.6\% for Meta Harness, and 58.7\% for the baseline). Our key observation was that optimization gains compounded only when regression control was built into the optimization loop, providing an inductive bias against shortcut solutions that fail to generalize.
\end{abstract}

\section{Introduction}

\subsection{Motivation}
A growing body of work optimizes agent harnesses (prompts, tools, memory, and even harness code) against a fixed set of benchmark tasks and reports the resulting improvement as evidence that the optimizer works. This is the setting in which most published agent-optimization results are obtained: an agent is optimized once, against a task set that does not change during or after optimization. Production agents, however, are rarely optimized only once. As new failures are observed in deployment, or as new task types are added to a product surface, an agent that has already been optimized once is optimized again, on top of whatever changes the first round of optimization already made. A one-shot benchmark score does not tell us whether an optimizer behaves well under this kind of repeated, recursive application.

\subsection{The Compounding Question}
We refer to this as the \emph{compounding question}: after an agent has been optimized once by a given method, can that method be applied again, on a task stream that has since expanded to include new tasks, without unwinding the gains the first round of optimization produced? An optimizer that compounds should (i) produce updates from the first round that generalize at least somewhat to tasks it did not see during search, and (ii) continue to make progress in a second round of optimization without regressing on tasks it had already solved. Neither property is guaranteed by strong performance on a static, single-phase benchmark, and to our knowledge no existing evaluation protocol for agent-harness optimizers directly measures both.

\subsection{Preview of Protocol and Findings}
\begin{figure}[t]
\centering
\begin{tabular}{@{}p{0.52\textwidth}@{\hspace{0.02\textwidth}}p{0.44\textwidth}@{}}
\strut\par\vspace{-\baselineskip}
\centering
\footnotesize
\setlength{\tabcolsep}{4pt}
\begin{tabular}{@{}l c cc c@{}}
\toprule
\multirow{2}{*}{Agent} & \multirow{2}{*}{Phase 1} & \multicolumn{2}{c}{Phase 2} & \multirow{2}{*}{Lifelong Avg.} \\
\cmidrule(lr){3-4}
& & Transfer & Re-opt. & \\
\midrule
Baseline & 62.5\% & 56.8\% & 56.8\% & 58.7\% \\
GEPA & 70.8\% & 54.5\% & 72.7\% & 66.0\% \\
Meta Harness & 66.6\% & 68.2\% & 59.1\% & 64.6\% \\
RELAI-VCL & \textbf{79.2\%} & \textbf{72.7\%} & \textbf{77.3\%} & \textbf{76.4\%} \\
\bottomrule
\end{tabular}
\captionof{table}{RELAI-VCL leads at every stage of the phased evaluation: the strongest static Phase 1 result, the only positive transfer to unseen tasks, continued gains after Phase 2 re-optimization, and the highest lifelong average pass rate. GEPA transfers below baseline (overfitting to Phase 1) and Meta Harness stalls under re-optimization.}
\label{tab:headline}
&
\strut\par\vspace{-\baselineskip}\vspace{-5pt}
\centering
\includegraphics[width=\linewidth]{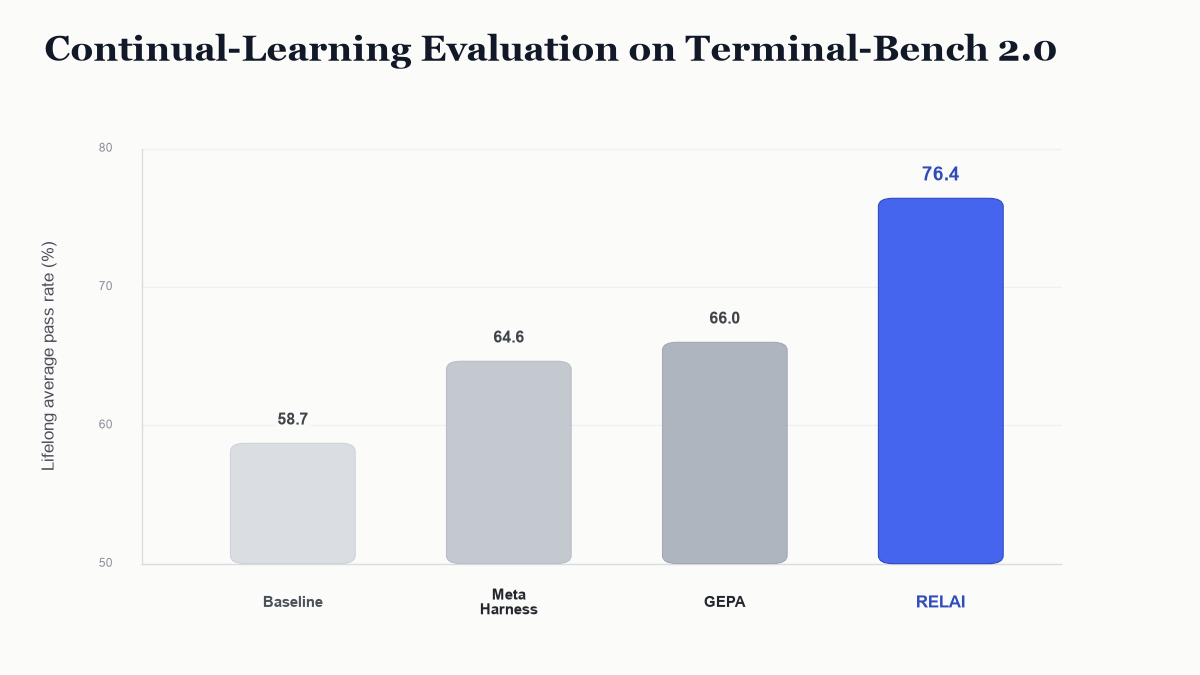}
\captionof{figure}{Lifelong average pass rate by agent: RELAI-VCL reaches 76.4\%, ahead of GEPA (66.0\%), Meta Harness (64.6\%), and the unoptimized baseline (58.7\%).}
\label{fig:lifelong-preview}
\end{tabular}
\end{figure}

We construct a two-phase continual-learning evaluation from hard tasks in Terminal-Bench 2.0. In Phase 1, each optimizer is given a fixed rollout budget to optimize a common baseline agent on an initial task set $\mathcal{T}_1$. We then evaluate the Phase-1-optimized agent both on $\mathcal{T}_1$ (the conventional static setting) and, without any further optimization, on an expanded task union $\mathcal{T}_1 \cup \mathcal{T}_2$ that includes a newly introduced task set $\mathcal{T}_2$ (a transfer setting). In Phase 2, each method receives an additional rollout budget, starting from its own Phase-1-optimized agent, to optimize on the combined task set $\mathcal{T}_1 \cup \mathcal{T}_2$, and we evaluate the resulting agent on the full union. This sequence lets us separate three distinct properties that a single benchmark score conflates: static optimization strength, transfer to unseen tasks, and the ability to keep improving once those tasks are incorporated into the search objective. Empirically, we find that all three methods improve over the baseline in the static Phase 1 setting, but the methods diverge sharply once task arrival and repeated optimization are introduced: only RELAI-VCL shows both positive transfer and continued improvement, while GEPA overfits to Phase 1 (transferring below baseline) and Meta Harness transfers well but stalls under re-optimization. Table~\ref{tab:headline} and Figure~\ref{fig:lifelong-preview} summarize these results; the full per-stage breakdown and analysis are given in Section~6.

\subsection{Contributions}
\begin{itemize}
    \item A phased continual-learning evaluation protocol built on Terminal-Bench 2.0 hard tasks.
    \item A head-to-head comparison of GEPA, Meta Harness, and RELAI-VCL under identical phased-optimization budgets.
    \item Evidence that regression control must live inside the search loop, not as post-hoc analysis, to obtain both transfer and continued improvement.
    \item Discussion of what a more realistic continual-learning benchmark for agents would need.
\end{itemize}

\section{Background and Related Work}

\subsection{Optimizing Agent Harnesses}
Recent agent-optimization methods treat the harness around a fixed LLM (its prompts, context, tool definitions, and control code) as a search space to be optimized against task feedback, rather than hand-tuned. GEPA optimizes agent prompts through reflective evolutionary search: it mutates candidates using natural-language reflection on rollout traces and retains candidates that improve measured performance~\citep{gepa2025}. MIPROv2 similarly searches over instructions and few-shot demonstrations for multi-stage LM programs using a surrogate-guided (Bayesian-optimization-style) procedure~\citep{miprov2}, and both build on DSPy, which compiles declarative LM pipelines into optimizable programs~\citep{dspy2023}. ACE instead treats an agent's context as an evolving playbook that accumulates and curates strategies over time rather than being rewritten from scratch, explicitly targeting the brevity bias and context collapse that repeated prompt rewriting can cause~\citep{ace2025}; Combee scales this style of prompt learning across many agents running in parallel~\citep{combee2026}. Reflexion attaches verbal, LLM-generated self-critique to a fixed agent architecture, using linguistic feedback in place of a scalar reward signal~\citep{reflexion2023}. Multi-Agent Design (MAAS) goes a step further and jointly searches prompts and agent topology, though over a restricted set of predefined block types~\citep{maas2025}. Maestro targets this same gap with a richer topology space, jointly searching non-sequential DAGs, persistent/global state, and context gating alongside node-level configuration, and is reported to outperform prompt-only optimizers such as GEPA and MIPROv2 even when restricted to prompt-only search, with further gains once graph search is enabled~\citep{maestro2025}. GRPO represents a different mechanism entirely: group-relative policy-gradient fine-tuning of model weights~\citep{grpo}, which changes the underlying model rather than the harness wrapped around it. Hyperagents pushes optimization one level further by folding the optimizer itself into a single self-modifiable program, so that the agent's own improvement process can also be revised, though its evaluation remains benchmark-bound~\citep{hyperagents2026}.

Closest to our setting is Meta-Harness, an outer-loop method that uses an agentic proposer (itself an LLM-driven coding agent with access to prior candidates' code, traces, and scores) to directly edit harness code and accept or reject the result based on evaluation feedback~\citep{metaharness2026}. Meta-Harness is one of the three optimizers we compare empirically in this paper (Section~4.3). What is common across this entire line of work is that a single, static evaluation set drives the search: no method in this cluster builds an explicit mechanism for detecting or preventing regressions on tasks outside that set into the search procedure itself, a gap we return to in Section~2.5.

\subsection{Continual Learning and Catastrophic Forgetting}
The compounding question we study is closely related to catastrophic forgetting in continual learning: the tendency of a model updated on new data to lose performance on previously learned tasks. The canonical regularization-based approach, Elastic Weight Consolidation, penalizes movement away from parameters that were important for earlier tasks, using a quadratic penalty derived from a Laplace approximation to the previous task's posterior~\citep{kirkpatrick2017ewc}. Regression-aware optimization, as used by RELAI-VCL and analyzed in this paper, plays an analogous role at the level of harness edits rather than model weights: rather than regularizing parameters, it constrains the search process itself to reject candidate harness edits that regress on tasks the agent could already solve.

A growing set of practical systems targets a related goal, durable agent improvement from experience, at layers below harness-code search. Memory- and skill-accumulation systems retrieve, consolidate, and write back information as the agent operates: examples include mem0~\citep{mem0} and Letta~\citep{letta} as general-purpose memory layers, Karpathy's structured-memory ``LLM wiki'' schema~\citep{karpathyllmwiki}, Anthropic's session-consolidation ``Dreaming'' feature~\citep{dreaming2026}, Nous Research's Hermes agent that autonomously writes and reorganizes its own skills~\citep{hermes2026}, and MetaClaw, which combines a skill bank with opportunistic reinforcement-learning fine-tuning~\citep{metaclaw2026}. A second group of systems closes the loop from production traces to harness code more directly: HALO diagnoses systemic failure modes from execution traces and feeds them to a coding agent~\citep{halo2026}, and LangSmith Engine clusters and diagnoses production traces into proposed evaluators and fixes~\citep{langsmithengine2026}; coding agents such as Codex~\citep{codex2025} and Claude Code~\citep{claudecode2025} are often used as the human-directed executors of the resulting edits, and production agent-serving systems such as LithosAI's Motus optimize harness, orchestration, and memory directly from production traces~\citep{lithosmotus2026}. Across this landscape, a proposed update is typically checked, if at all, only \emph{after} it has already been made: a held-out evaluation, a shadow deployment, or a human reviewing a pull request. None of these mechanisms make regression-avoidance a constraint that shapes the search or update process while it is running, which is exactly the property Section~2.5 and Section~4.4 identify as distinguishing RELAI-VCL, and the property this paper tests empirically.

\subsection{Benchmark Overfitting and the Static-Benchmark Critique}
A single post-optimization benchmark score is a weak signal of generalization: an optimizer with a large enough search budget and a fixed, finite evaluation set can find candidates that improve the measured score without improving (or while actively harming) performance on tasks outside that set. This is a specific instance of the broader concern that optimizing against a fixed proxy metric can produce solutions that satisfy the metric without satisfying the underlying goal it was meant to approximate, a failure mode with several distinct mechanisms in the Goodhart's-Law literature~\citep{manheim2018goodhart}. A concrete precedent for this exact failure mode in language models is IFBench, which finds that models trained to follow a small set of verifiable instruction-following constraints overfit to those constraints and generalize poorly to new, unseen ones~\citep{ifbench}. Our two-phase protocol is designed specifically to surface the analogous failure mode in agent-harness optimization: an optimizer that overfits Phase 1 should show strong Phase 1 gains but negative or weak transfer once $\mathcal{T}_2$ is introduced, which is exactly the pattern we observe for GEPA in Section~6.

\subsection{Terminal-Bench as an Evaluation Substrate}
Terminal-Bench 2.0 provides a set of terminal-based agent tasks with automatically checkable success criteria, spanning domains such as scientific computing, software engineering, machine-learning infrastructure, systems administration, and security-relevant workflows~\citep{terminalbench2026}. Its verifiable, per-task pass/fail outcomes make it well suited to the phased protocol in Section~3, which requires repeatedly re-evaluating the same and newly introduced tasks under different optimized agents. At the same time, as we discuss in Section~7, Terminal-Bench's tasks are largely independent of one another, which is a simplifying assumption relative to production settings where failures are often correlated across a shared application domain.

\subsection{RELAI-VCL}
We describe RELAI-VCL in the same technical register as the preceding subsections: it is a regression-aware continual-learning optimizer for agent harnesses, which we analyze here as a method under evaluation rather than as a product. Mechanically, it can be viewed as an instance of constrained candidate search: rather than optimizing pass rate on the current task set unconditionally, it rejects any candidate harness edit that improves performance on newly targeted tasks at the cost of regressing on tasks the agent previously solved. Sections~2.1 and 2.2 surveyed two distinct clusters of prior work (static-benchmark harness/prompt optimizers, and practical memory-, skill-, and trace-to-harness systems), and in both, validation of a proposed update happens after the update is made, if it happens at all. RELAI-VCL's distinguishing property, and the one this paper evaluates empirically in Sections~5-6, is enforcing the no-regression constraint \emph{inside} the search loop rather than checking for regressions only after search has converged.

\section{Continual-Learning Evaluation Protocol}
\subsection{Problem Setup}
We evaluate agent optimizers in a phased setting intended to capture a minimal form of continual learning. Let $\mathcal{T}_1$ denote an initial set of tasks and $\mathcal{T}_2$ denote a later set of newly introduced tasks. The task stream therefore expands from $\mathcal{T}_1$ to $\mathcal{T}_1 \cup \mathcal{T}_2$. Let $A_0$ denote the fixed baseline agent before optimization.

\subsection{Phased Optimization Protocol}
Starting from $A_0$, an optimizer first receives an optimization budget on $\mathcal{T}_1$, producing an updated agent $A_1$. The resulting agent is then evaluated not only on the original tasks but also on the expanded task union $\mathcal{T}_1 \cup \mathcal{T}_2$, before any additional optimization on the new tasks. This transfer evaluation measures whether the first optimization round produced changes that generalize beyond the tasks that directly drove the search.

The second phase evaluates whether optimization can continue after the first update has already modified the agent. Starting from $A_1$, the optimizer receives an additional budget on the combined task set $\mathcal{T}_1 \cup \mathcal{T}_2$, producing $A_2$. The final evaluation measures pass rate on the full task union. This phase tests a stronger property than transfer alone: the optimizer must improve on the expanded task set while retaining useful behavior acquired during the first phase.

\subsection{Metrics}
The primary metric is pass rate on a specified task set. We report: (i) Phase 1 pass rate after optimization on $\mathcal{T}_1$; (ii) transfer pass rate on $\mathcal{T}_1 \cup \mathcal{T}_2$ after optimizing only on $\mathcal{T}_1$; (iii) final pass rate on $\mathcal{T}_1 \cup \mathcal{T}_2$ after the second optimization phase; and (iv) a lifelong average pass rate aggregating performance across the staged evaluation.

Every task is evaluated twice per agent, and a task is scored as the fraction of its two trials that the agent solves (0, $\tfrac{1}{2}$, or 1). Formally, writing $R = 2$ for the number of repeated trials per task and $\mathbbm{1}[A \text{ solves } t \text{ on trial } r]$ for the outcome of trial $r$ of task $t$, the pass rate of agent $A$ on task set $\mathcal{T}$ is
\[
\text{PassRate}(A, \mathcal{T}) = \frac{1}{R\,|\mathcal{T}|}\sum_{t \in \mathcal{T}} \sum_{r=1}^{R} \mathbbm{1}[A \text{ solves } t \text{ on trial } r],
\]
and
\begin{align*}
\text{Phase-1} &= \text{PassRate}(A_1, \mathcal{T}_1) \\
\text{Transfer} &= \text{PassRate}(A_1, \mathcal{T}_1 \cup \mathcal{T}_2) \\
\text{Final} &= \text{PassRate}(A_2, \mathcal{T}_1 \cup \mathcal{T}_2) \\
\text{LifelongAvg} &= \tfrac{1}{3}\left(\text{Phase-1} + \text{Transfer} + \text{Final}\right)
\end{align*}
The LifelongAvg definition above is an unweighted mean of the three reported pass rates; we verified it reproduces every lifelong-average number reported in Section~6 exactly (e.g., for the baseline, $\tfrac{1}{3}(62.5\% + 56.8\% + 56.8\%) = 58.7\%$, since the unoptimized baseline's ``Final'' score is identical to its ``Transfer'' score). For the baseline agent, which is never re-optimized, Transfer and Final therefore coincide by construction.

\subsection{Why Static Evaluation Is Insufficient}
This protocol differs from static benchmark optimization. A one-shot benchmark score measures whether an optimizer can improve an agent on the tasks it is allowed to optimize against. The phased protocol additionally asks whether those improvements survive task expansion and whether subsequent optimization compounds with earlier updates rather than replacing or undoing them. These properties are central for deployed agents, where optimization is expected to run recursively as new failures are observed.

\section{Methods Compared}
\subsection{Baseline Agent}
All methods start from the same fixed baseline terminal agent using GPT-5.5 as the underlying LLM. This baseline agent is never re-optimized itself; it serves as the common starting point $A_0$ from which every optimizer produces its own $A_1$ and $A_2$: GEPA, Meta Harness, and RELAI-VCL each optimize their own private copy of this identical starting agent, rather than sharing one mutable copy. The agent is built on Harbor~\citep{Harbor_Framework}, an open-source framework for evaluating agents and the official framework for Terminal-Bench 2.0 evaluation, and extends Harbor's Terminus2 base agent with native LLM tool calling in place of Terminus2's default in-context JSON/XML action parsing. Its action space consists of three tools: \texttt{execute\_commands} (batches of tmux keystrokes, each paired with a required analysis-and-plan rationale), \texttt{task\_complete} (gated behind a checklist prompt that asks the agent to re-verify minimal state changes before finishing), and \texttt{image\_read} (reads an image file from the container and sends it to the model multimodally, since the agent otherwise has no way to inspect non-text files). The harness components exposed to each optimizer's search space are the system prompt, the tool definitions, and the surrounding control code; full implementation details are given in Appendix~B.

\subsection{GEPA}
GEPA is evaluated as an evolutionary optimization method over the agent harness, using reflective mutation guided by natural-language feedback on rollout traces. We evaluate two variants: a prompt-only variant, which restricts the search space to the agent's prompts, and a code-mutation variant (``GEPA-code''), which additionally allows edits to harness code. GPT-5.5 is used as GEPA's proposer model in both variants. In the experiments summarized here, GEPA-code failed to produce a valid candidate during Phase 1 and is therefore excluded from the Phase 2 comparisons; only GEPA's prompt-only variant is carried forward. As discussed in Section~6.1, closer inspection of GEPA's Phase-1-optimized prompt revealed sample-specific information that appeared to be hardcoded from the Phase 1 tasks, consistent with the negative transfer result in Section~6.2. Concretely, GEPA's prompt grows from 5 lines before optimization to 103 lines after Phase 1 and 195 lines after Phase 2, with the added text organized as named, per-task lessons; Appendix~C gives the full breakdown.

\subsection{Meta Harness}
Meta Harness is evaluated as an outer-loop harness-code search method: an agentic proposer directly edits the harness implementation, and candidates are accepted or rejected based on evaluation feedback on the current task set. The proposer agent is Codex, itself powered by GPT-5.5. Unlike GEPA, Meta Harness's search space targets harness code changes directly rather than prompt text. Its two accepted candidates were both general-purpose harness robustness fixes rather than task-specific changes: the Phase 1 candidate hardens command-completion detection and malformed tool-call handling, and the Phase 2 candidate adds a pass that compacts long, repetitive terminal output (package-manager logs, build/download progress, test-runner boilerplate) into short summaries; Appendix~C gives further detail.

\subsection{RELAI-VCL}
RELAI-VCL (RELAI's Verifiable Continual Learning) is evaluated as a regression-aware continual-learning optimizer. It searches over the same broad space of possible harness edits as the other methods (prompt changes, tool patches, workflow modifications, memory corrections, skills, or code edits) but incorporates a no-regression constraint directly into candidate selection: any candidate that improves performance on newly targeted tasks by sacrificing previously working behavior is rejected during search, rather than being accepted and only flagged afterward. The intended effect is to bias search toward the smallest intervention that resolves a given failure without disrupting existing behavior. In these experiments, RELAI-VCL uses GPT-5.5 as its proposer model. Its accepted candidates add a general task-type classifier (distinguishing code-editing from operational/service tasks) and a completion gate that runs verifier-discovery and runtime-contract checks before accepting task completion, rather than task-specific fixes; Appendix~C gives further detail.

\subsection{Method Comparison Summary}
The proposer model listed for each optimizer below is the specific configuration used in this evaluation, not an inherent property of the method itself; a different proposer model could in principle be substituted for any of them. Table~\ref{tab:method-comparison} summarizes the four agents compared in this evaluation.

\begin{table}[H]
\centering
\caption{Comparison of the baseline agent and the three optimizers evaluated in this paper. All methods share the same baseline agent, LLM, and per-phase rollout budget; they differ in search space, proposer model, and whether regression control is enforced during search or only measured afterward. \emph{Post-hoc only} means regressions on prior tasks are, at best, checked after a candidate is already accepted, rather than shaping the search itself; \emph{enforced in-loop} means candidates that regress on prior tasks are rejected during search. GEPA's code-mutation variant additionally failed to produce a valid candidate and was excluded from Phase 2.}
\label{tab:method-comparison}
\begin{tabularx}{\textwidth}{@{}l X X X X@{}}
\toprule
Method & Search space & Proposer model & Regression handling & Rollout budget/phase \\
\midrule
Baseline & --- (fixed) & --- & --- & --- \\
GEPA (prompt) & Prompts & GPT-5.5 & Post-hoc only & 200 \\
GEPA (code) & Prompts + harness code & GPT-5.5 & Post-hoc only & 200 \\
Meta Harness & Harness code & Codex (GPT-5.5) & Post-hoc only & 200 \\
RELAI-VCL & Prompt, tool, workflow, memory, skill, code & GPT-5.5 & Enforced in-loop & 200 \\
\bottomrule
\end{tabularx}
\end{table}

Low-level implementation details, prompts, and code diffs for all methods are deferred to Appendix~C.

\section{Experimental Setup}
\subsection{Terminal-Bench 2.0 Task Split}
We instantiate the phased protocol using hard tasks from Terminal-Bench 2.0. Phase 1 consists of 12 tasks with an agent timeout of at most 900 seconds. Phase 2 introduces 10 additional tasks with an agent timeout of at most 1800 seconds. The resulting task union contains 22 tasks spanning terminal-based problem solving settings such as systems work, machine-learning infrastructure, and security-relevant workflows.

Table~\ref{tab:task-split} summarizes the two phases. Both phases are drawn from the same difficulty tier: Terminal-Bench 2.0 annotates each of its 89 tasks with a difficulty level and an agent timeout, and 30 of the 89 tasks are annotated \emph{hard}. $\mathcal{T}_1$ is exactly the 12 hard tasks with a 900-second agent timeout, and $\mathcal{T}_2$ is exactly the 10 additional hard tasks with a 1800-second agent timeout; the remaining 8 hard tasks, which have a 3600- or 7200-second agent timeout, are excluded from both phases. The full task IDs are given in Appendix~A.

\begin{table}[H]
\centering
\caption{Terminal-Bench 2.0 task split used for the phased protocol.}
\label{tab:task-split}
\begin{tabular}{@{}lccl@{}}
\toprule
Task set & \# Tasks & Max.\ agent timeout & Difficulty \\
\midrule
$\mathcal{T}_1$ (Phase 1) & 12 & 900s & Hard \\
$\mathcal{T}_2$ (Phase 2) & 10 & 1800s & Hard \\
$\mathcal{T}_1 \cup \mathcal{T}_2$ & 22 & 900s / 1800s & Hard \\
\bottomrule
\end{tabular}
\end{table}

\subsection{Optimization Budget}
In Phase 1, each optimizer receives 200 rollouts to optimize the agent harness against the 12 Phase 1 tasks. In Phase 2, each method receives another 200 rollouts starting from its Phase 1-optimized agent and optimizing on the combined Phase 1 and Phase 2 task set. The budget is identical across methods to keep the comparison controlled.

\subsection{Evaluation Procedure}
After Phase 1 optimization, we evaluate each Phase 1-optimized agent on the 12 Phase 1 tasks (the Phase 1 metric of Section~3.3). We then evaluate the same Phase 1-optimized agent on the combined 22-task union without additional optimization, measuring transfer to the newly introduced Phase 2 tasks (the Transfer metric). After Phase 2 optimization, the final agent is evaluated on the full task union (the Final metric).

\section{Results}
\subsection{Phase 1: Static Optimization}
\begin{figure}[H]
\centering
\begin{tabular}{@{}m{0.56\textwidth}@{\hspace{0.02\textwidth}}m{0.38\textwidth}@{}}
\centering
\includegraphics[width=\linewidth]{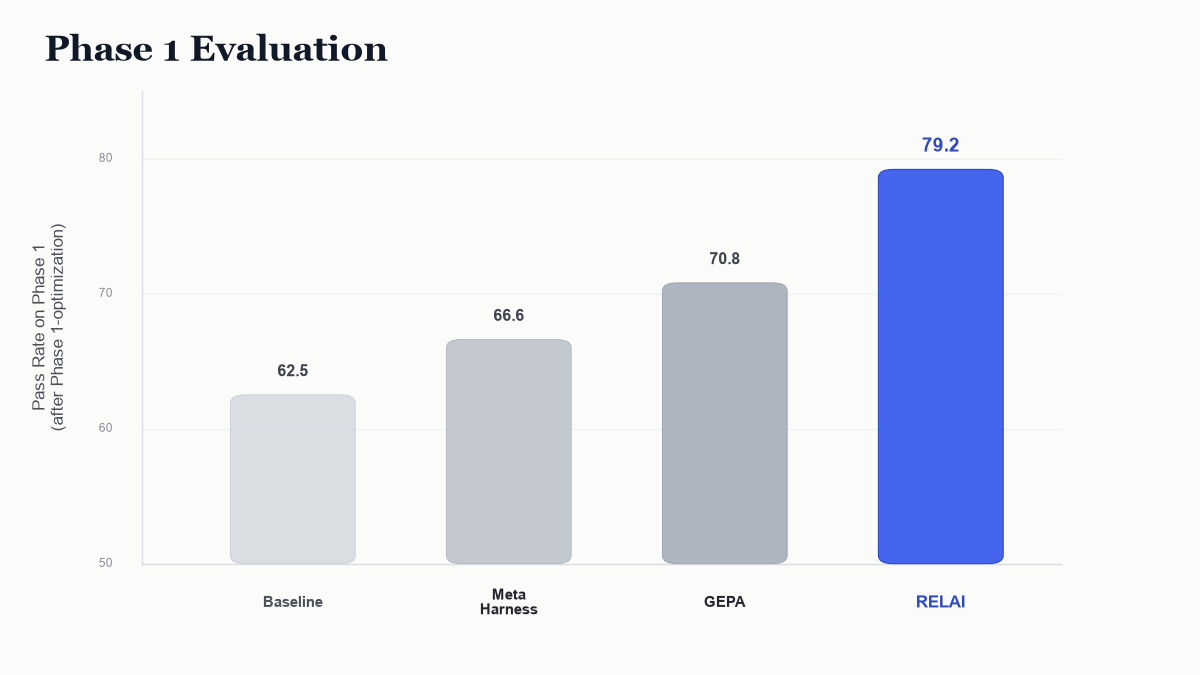}
\captionof{figure}{Pass rate on the 12 Phase 1 tasks after Phase 1 optimization.}
\label{fig:phase1}
&
\centering
\footnotesize
\begin{tabular}{@{}l>{\centering\arraybackslash}p{2.1cm}@{}}
\toprule
Agent & Phase 1 pass rate \\
\midrule
Baseline & 62.5\% \\
GEPA & 70.8\% \\
Meta Harness & 66.6\% \\
RELAI-VCL & \textbf{79.2\%} \\
\bottomrule
\end{tabular}
\captionof{table}{Phase 1 pass rate (static optimization) on $\mathcal{T}_1$ (12 tasks).}
\label{tab:phase1}
\end{tabular}
\end{figure}

On the 12 Phase 1 tasks, the baseline pass rate is 62.5\%, GEPA reaches 70.8\%, Meta Harness reaches 66.6\%, and RELAI-VCL reaches 79.2\% (Figure~\ref{fig:phase1}, Table~\ref{tab:phase1}). This stage corresponds most closely to the conventional static benchmark setting: each optimizer is evaluated on the same task set used during the first optimization phase.

RELAI-VCL achieves the strongest Phase 1 result: a $+16.7$-point gain over the baseline and a $+8.4$-point gain over the best alternative, GEPA's prompt optimizer. Beyond the aggregate score, the harness changes RELAI-VCL applied appeared targeted and broadly generalizable rather than encoding task-specific shortcuts: its accepted candidates add a general task-type classifier and a completion gate that runs verifier-discovery and runtime-contract checks, with no task IDs or literal expected outputs written into the prompt template itself (Appendix~C). GEPA's prompt optimizer improves the baseline by approximately 8 points, but inspection of the resulting prompt revealed sample-specific information that appeared to be hardcoded from the Phase 1 tasks, suggesting overfitting to the Phase 1 task set specifically, a pattern whose consequences become visible in the transfer results below. Meta Harness improves more modestly, by approximately 4 points, with changes that appeared generic and in some cases conservative; this limits its Phase 1 gain but turns out to matter for how well it transfers.

\subsection{Transfer to Newly Introduced Tasks}
\begin{figure}[H]
\centering
\begin{tabular}{@{}m{0.56\textwidth}@{\hspace{0.02\textwidth}}m{0.38\textwidth}@{}}
\centering
\includegraphics[width=\linewidth]{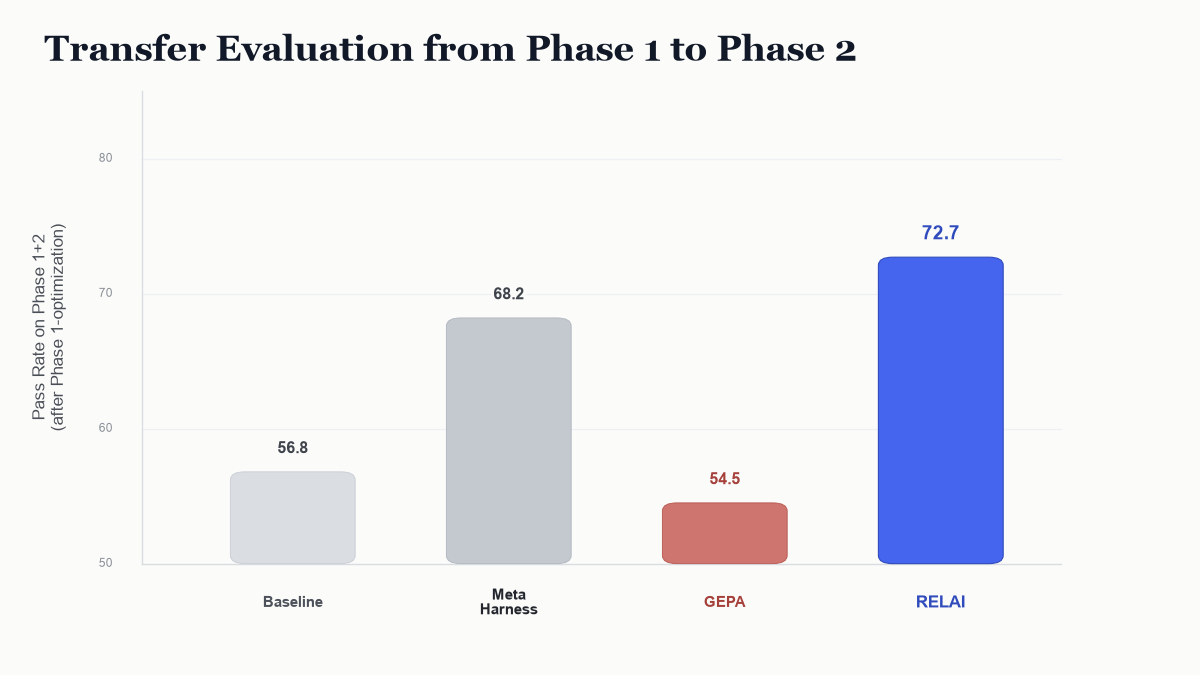}
\captionof{figure}{Pass rate on the 22-task union $\mathcal{T}_1 \cup \mathcal{T}_2$ after Phase 1 optimization only (transfer setting).}
\label{fig:transfer}
&
\centering
\footnotesize
\begin{tabular}{@{}l>{\centering\arraybackslash}p{2.1cm}@{}}
\toprule
Agent & Transfer pass rate \\
\midrule
Baseline & 56.8\% \\
GEPA & 54.5\% \\
Meta Harness & 68.2\% \\
RELAI-VCL & \textbf{72.7\%} \\
\bottomrule
\end{tabular}
\captionof{table}{Transfer pass rate on $\mathcal{T}_1 \cup \mathcal{T}_2$ (22 tasks) after Phase 1 optimization only.}
\label{tab:transfer}
\end{tabular}
\end{figure}

We next evaluate each Phase-1-optimized agent on the combined 22-task union $\mathcal{T}_1 \cup \mathcal{T}_2$, without any further optimization (Figure~\ref{fig:transfer}, Table~\ref{tab:transfer}). This measures whether the first optimization round produced changes that generalize to tasks that did not drive the search. The baseline pass rate is 56.8\%, GEPA reaches 54.5\%, Meta Harness reaches 68.2\%, and RELAI-VCL reaches 72.7\%.

GEPA transfers negatively, falling from 70.8\% on $\mathcal{T}_1$ to 54.5\% on the union, below the 56.8\% unoptimized baseline. This is consistent with the sample-specific prompt changes noted in Section~6.1: gains on the Phase 1 tasks failed to generalize once the task distribution expanded, providing evidence that GEPA overfit to the Phase 1 benchmark specifically. Meta Harness transfers much better, reaching 68.2\% on the combined set: its comparatively generic, conservative Phase 1 updates generalized well beyond the tasks used during optimization. RELAI-VCL achieves the strongest transfer result at 72.7\%; its pass rate on the unseen Phase 2 tasks alone is 65.0\%, about 15 points above the baseline despite never having been optimized against those tasks. Taken together, these results show that static Phase 1 gains alone do not predict how an optimized agent performs once new tasks arrive.

\subsection{Phase 2: Re-optimization on the Combined Task Set}
\begin{figure}[H]
\centering
\begin{tabular}{@{}m{0.56\textwidth}@{\hspace{0.02\textwidth}}m{0.38\textwidth}@{}}
\centering
\includegraphics[width=\linewidth]{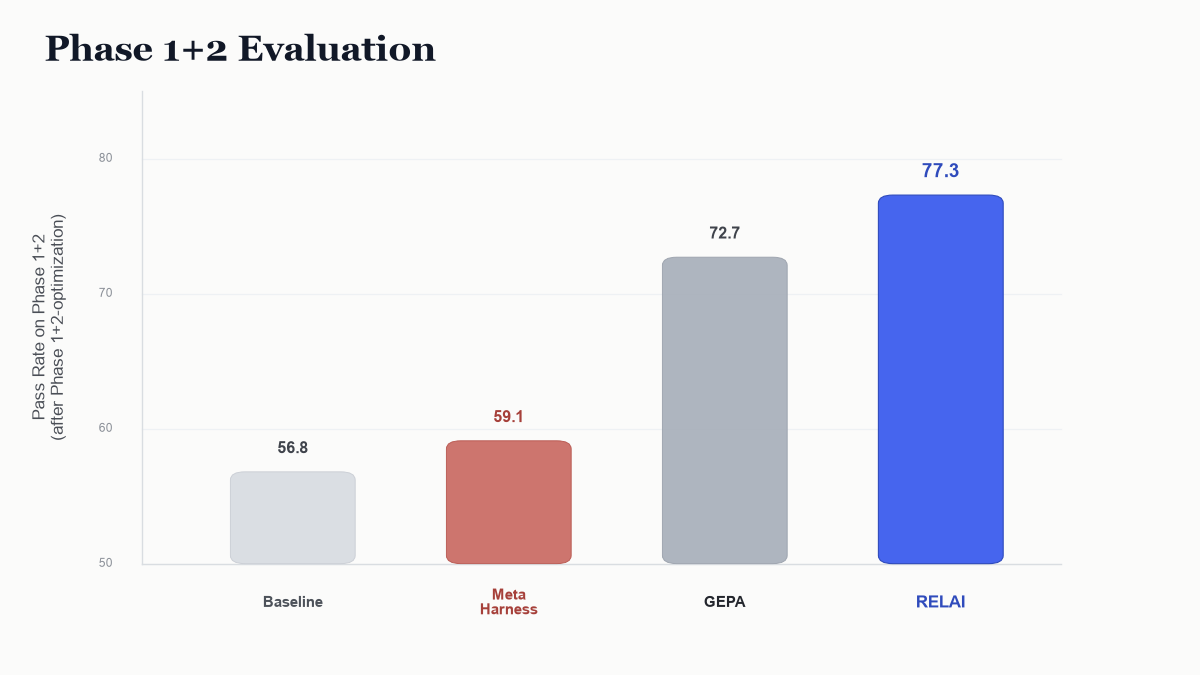}
\captionof{figure}{Pass rate on the 22-task union $\mathcal{T}_1 \cup \mathcal{T}_2$ after Phase 2 re-optimization on the combined task set.}
\label{fig:phase2}
&
\centering
\footnotesize
\begin{tabular}{@{}l>{\centering\arraybackslash}p{2.1cm}@{}}
\toprule
Agent & Final pass rate \\
\midrule
Baseline (not re-opt.) & 56.8\% \\
GEPA & 72.7\% \\
Meta Harness & 59.1\% \\
RELAI-VCL & \textbf{77.3\%} \\
\bottomrule
\end{tabular}
\captionof{table}{Final pass rate on $\mathcal{T}_1 \cup \mathcal{T}_2$ (22 tasks) after Phase 2 re-optimization.}
\label{tab:phase2}
\end{tabular}
\end{figure}

We next give each method a second, identical 200-rollout budget, starting from its own Phase-1-optimized agent, to optimize on the combined task set $\mathcal{T}_1 \cup \mathcal{T}_2$ (Figure~\ref{fig:phase2}, Table~\ref{tab:phase2}). This tests whether an optimizer can continue improving once its first round of updates is already in place. GEPA reaches 72.7\%, Meta Harness reaches 59.1\%, and RELAI-VCL reaches 77.3\%.

GEPA recovers to 72.7\% once the Phase 2 tasks are included in its optimization objective, following the same pattern as in Phase 1: it performs well once the relevant tasks are directly optimized against, but its earlier negative transfer (Section~6.2) raises the question of how these Phase-2-conditioned gains would hold up under a further distribution shift. Meta Harness shows the opposite limitation: its Phase 1 changes transferred well, but it could not build on them further. Every candidate generated during the second optimization round performed worse than the existing agent, and its score \emph{fell} to 59.1\%. RELAI-VCL continues improving, reaching 77.3\%, the strongest result on the combined task set. It is therefore the only method in this evaluation that demonstrates both properties required for optimization gains to compound: strong transfer to unseen tasks (Section~6.2), and continued improvement once those tasks are incorporated into the search objective. Transfer alone is not sufficient if the optimizer subsequently stalls, as Meta Harness illustrates; repeated optimization alone is not sufficient if each round overfits to the tasks currently being evaluated, as GEPA illustrates. Compounding, in this evaluation, requires both.

\subsection{Lifelong Average Performance}
\begin{figure}[H]
\centering
\begin{tabular}{@{}m{0.56\textwidth}@{\hspace{0.02\textwidth}}m{0.38\textwidth}@{}}
\centering
\includegraphics[width=\linewidth]{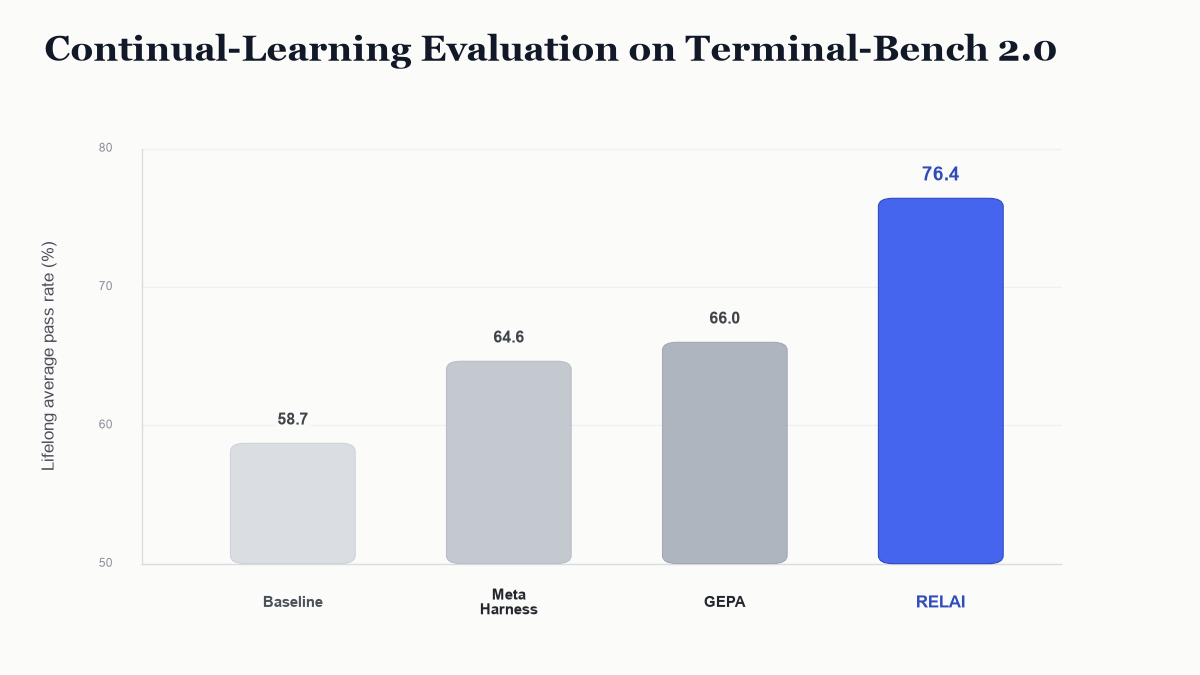}
\captionof{figure}{Lifelong average pass rate: the mean of the Phase-1, Transfer, and Final pass rates for each agent.}
\label{fig:lifelong}
&
\centering
\footnotesize
\begin{tabular}{@{}l>{\centering\arraybackslash}p{2.1cm}@{}}
\toprule
Agent & Lifelong avg.\ pass rate \\
\midrule
Baseline & 58.7\% \\
GEPA & 66.0\% \\
Meta Harness & 64.6\% \\
RELAI-VCL & \textbf{76.4\%} \\
\bottomrule
\end{tabular}
\captionof{table}{Lifelong average pass rate per agent.}
\label{tab:lifelong}
\end{tabular}
\end{figure}

Figure~\ref{fig:lifelong} and Table~\ref{tab:lifelong} report the lifelong average pass rate defined in Section~3.3 (the unweighted mean of the Phase-1, Transfer, and Final pass rates), which aggregates static optimization strength, transfer, and continued-improvement behavior into a single number. The reported lifelong average pass rates are 58.7\% for the baseline, 66.0\% for GEPA, 64.6\% for Meta Harness, and 76.4\% for RELAI-VCL.

A single static score cannot distinguish an optimizer that overfits Phase 1 (GEPA) from one that transfers well but stalls (Meta Harness) from one that does both well (RELAI-VCL): GEPA and Meta Harness post similar lifelong averages (66.0\% vs.\ 64.6\%) despite failing in opposite ways, and both trail RELAI-VCL by more than 10 points. The lifelong average is only informative once it is decomposed into its three components (Sections~6.1-6.3); we report it here as a single-number summary, not as a replacement for the per-stage breakdown.

\subsection{Synthesis}
Table~\ref{tab:synthesis} consolidates every number reported in Sections~6.1-6.4 into a single view.

\begin{table}[H]
\centering
\caption{Consolidated results across all four evaluation stages.}
\label{tab:synthesis}
\begin{tabular}{@{}lcccc@{}}
\toprule
Agent & Phase 1 & Transfer & Final & Lifelong avg. \\
\midrule
Baseline & 62.5\% & 56.8\% & 56.8\% & 58.7\% \\
GEPA & 70.8\% & 54.5\% & 72.7\% & 66.0\% \\
Meta Harness & 66.6\% & 68.2\% & 59.1\% & 64.6\% \\
RELAI-VCL & \textbf{79.2\%} & \textbf{72.7\%} & \textbf{77.3\%} & \textbf{76.4\%} \\
\bottomrule
\end{tabular}
\end{table}

RELAI-VCL is the only method that leads at every one of the four evaluated stages.

\subsection{Interpretation: Regression Control as Generalization Filter}
The pattern across Sections~6.1-6.5 is consistent with regression-aware search serving two purposes simultaneously. First, it directly targets forgetting: by construction, RELAI-VCL rejects any candidate that regresses on tasks the agent could already solve, so its Phase 2 update cannot trade Phase 1 competence for Phase 2 gains in the way GEPA's Phase 1 update appears to have traded general competence for Phase-1-task-specific competence. Second, and more speculatively, the no-regression constraint appears to act as an implicit generalization filter: an update that improves a new task \emph{without} breaking any previously solved task is, in this task population, less likely to be a narrow, task-specific shortcut and more likely to address a more fundamental aspect of agent behavior (Section~2.2, Section~2.5). This connects the empirical pattern in Section~6.2 (RELAI-VCL's Phase 1 updates transferring positively to entirely unseen tasks) to the mechanism described in Section~4.4, rather than treating it as a coincidence. We state this as an interpretation consistent with the evidence in this evaluation, not as a general claim about regression-aware optimization; Section~7 discusses why the current protocol cannot fully distinguish ``generalization filter'' from other explanations.

\section{Limitations and Toward Realistic Continual-Learning Benchmarks}
The evaluation in this paper is a step toward more realistic continual-learning benchmarks for agents, not a complete one. Several gaps remain.

\subsection{Loosely Related Tasks}
The tasks in Terminal-Bench 2.0 are only loosely related to one another: $\mathcal{T}_1$ and $\mathcal{T}_2$ span distinct domains (systems work, ML infrastructure, security) with little shared structure between individual tasks. In practical deployments, failures usually arise within the same application domain and often involve shared tools, workflows, data sources, and policies. In such correlated-failure settings, optimizers may exploit cross-task correlations to discover domain-specific shortcuts, which could make both overfitting and generalization failures more pronounced than what we observe here, in either direction, for any of the compared methods.

\subsection{Repeatability and Verifier Availability}
Methods such as GEPA and Meta Harness (and our own protocol) assume that tasks can be executed repeatedly and that reliable evaluators or verifiers are already available to score each rollout. In practice this is rarely true: production feedback is often a single observed agent trajectory, accompanied by incomplete information about what went wrong, with no direct way to reproduce the failure on demand or to verify that a proposed fix actually resolves it.

\subsection{Toward More Realistic Benchmarks}
More realistic continual-learning benchmarks for agents should evaluate optimization under these harder constraints directly: task streams with realistic cross-task correlation rather than loosely related tasks, limited observability (a single trajectory rather than a repeatable environment), incomplete feedback about failure causes, imperfect or partial verifiers, and long-horizon evaluation with many repeated rounds of optimization rather than the two phases studied here. These conditions remain significantly understudied in the agent-optimization literature, and we believe they deserve more attention from the community. 

\section{Conclusion}
This paper asked whether gains from agent-harness optimizers compound: after an agent has been optimized once, can it be optimized again on newly arrived tasks without losing the gains the first round produced? We answered this empirically with a two-phase continual-learning evaluation on Terminal-Bench 2.0, comparing GEPA, Meta Harness, and RELAI-VCL under identical optimization budgets. All three methods improved over the baseline agent in the static, single-phase setting, but only RELAI-VCL showed both positive transfer to unseen tasks and continued improvement once those tasks were incorporated into a second round of optimization, leading at every evaluated stage and on the lifelong average pass rate (76.4\% vs.\ 66.0\% for GEPA, 64.6\% for Meta Harness, and 58.7\% for the baseline). The measured takeaway from this evaluation is that regression-aware optimization, rejecting candidates that trade new-task gains for regressions on prior tasks, was associated with both transfer and continued improvement when enforced inside the optimization loop, rather than checked only after the fact. We report this as an evidence-based conclusion tied to this specific protocol and task population, not as a general law about agent optimizers; Section~7 outlines the harder, more realistic conditions under which this question should next be studied.

\setcitestyle{numbers}
\bibliographystyle{abbrvnat}
\bibliography{refs}

\newpage
\appendix

\section{Task List and Split Criteria}
Terminal-Bench 2.0 annotates each of its 89 tasks with a difficulty level (easy, medium, or hard) and an agent timeout. 30 of the 89 tasks are annotated hard. $\mathcal{T}_1$ is exactly the 12 hard tasks with a 900-second agent timeout, and $\mathcal{T}_2$ is exactly the 10 additional hard tasks with a 1800-second agent timeout. The remaining 8 hard tasks (\texttt{bn-fit-modify}, \texttt{circuit-fibsqrt}, \texttt{fix-ocaml-gc}, \texttt{install-windows-3.11}, \texttt{regex-chess}, \texttt{train-fasttext}, and \texttt{video-processing}, each with a 3600-second agent timeout, and \texttt{sam-cell-seg}, with a 7200-second agent timeout) are excluded from both phases. Table~\ref{tab:task-list} gives the full task IDs.

Qualitatively, the 22 tasks span a range of topics, including systems programming, cryptography, machine-learning infrastructure, computer graphics, and bioinformatics; this description is descriptive rather than a rigorous per-task category system, since Terminal-Bench 2.0 does not itself annotate a topic label per task.

\begin{table}[H]
\centering
\caption{Full task list for $\mathcal{T}_1$ and $\mathcal{T}_2$. All 22 tasks are annotated hard difficulty in Terminal-Bench 2.0.}
\label{tab:task-list}
\begin{tabular}{@{}llc@{}}
\toprule
Phase & Task ID & Agent timeout \\
\midrule
$\mathcal{T}_1$ & \texttt{cancel-async-tasks} & 900s \\
$\mathcal{T}_1$ & \texttt{configure-git-webserver} & 900s \\
$\mathcal{T}_1$ & \texttt{fix-code-vulnerability} & 900s \\
$\mathcal{T}_1$ & \texttt{gpt2-codegolf} & 900s \\
$\mathcal{T}_1$ & \texttt{make-doom-for-mips} & 900s \\
$\mathcal{T}_1$ & \texttt{model-extraction-relu-logits} & 900s \\
$\mathcal{T}_1$ & \texttt{password-recovery} & 900s \\
$\mathcal{T}_1$ & \texttt{polyglot-rust-c} & 900s \\
$\mathcal{T}_1$ & \texttt{sparql-university} & 900s \\
$\mathcal{T}_1$ & \texttt{torch-pipeline-parallelism} & 900s \\
$\mathcal{T}_1$ & \texttt{torch-tensor-parallelism} & 900s \\
$\mathcal{T}_1$ & \texttt{write-compressor} & 900s \\
\midrule
$\mathcal{T}_2$ & \texttt{dna-assembly} & 1800s \\
$\mathcal{T}_2$ & \texttt{extract-moves-from-video} & 1800s \\
$\mathcal{T}_2$ & \texttt{feal-differential-cryptanalysis} & 1800s \\
$\mathcal{T}_2$ & \texttt{feal-linear-cryptanalysis} & 1800s \\
$\mathcal{T}_2$ & \texttt{llm-inference-batching-scheduler} & 1800s \\
$\mathcal{T}_2$ & \texttt{make-mips-interpreter} & 1800s \\
$\mathcal{T}_2$ & \texttt{mcmc-sampling-stan} & 1800s \\
$\mathcal{T}_2$ & \texttt{path-tracing} & 1800s \\
$\mathcal{T}_2$ & \texttt{path-tracing-reverse} & 1800s \\
$\mathcal{T}_2$ & \texttt{protein-assembly} & 1800s \\
\bottomrule
\end{tabular}
\end{table}

\section{Harness and Agent Configurations}

\subsection{Baseline Agent}
The baseline agent (``TerminusKira'') is a Harbor-compatible wrapper around a subclass of Harbor's \texttt{Terminus2} agent that replaces its default in-context JSON/XML action parsing with native LLM tool calling via the \texttt{tools} parameter of the completion API. All API calls go through \texttt{litellm}, with \texttt{AGENT\_MODEL = "openai/gpt-5.5"} shared by all three optimizers' copies of the agent.

\paragraph{Tool definitions.} Three tools are exposed to the model:
\begin{itemize}
    \item \texttt{execute\_commands}: takes required \texttt{analysis} and \texttt{plan} string fields plus a \texttt{commands} array of \texttt{\{keystrokes, duration\}} objects, sent as literal tmux keystrokes. Command execution uses marker-based polling (an echo marker is appended after each command) so the harness can detect completion and move on before the requested duration elapses, rather than always waiting the full duration.
    \item \texttt{task\_complete}: takes no arguments. The first call is intercepted by a checklist prompt that asks the agent to re-read the task instructions, identify the minimal set of files that must change, and confirm no other side effects were introduced, before a second call actually ends the episode.
    \item \texttt{image\_read}: takes \texttt{file\_path} and \texttt{image\_read\_instruction}. The harness reads the file from the container as base64 and sends it as a multimodal user message so the agent can visually inspect image files it otherwise cannot read via shell commands.
\end{itemize}

\paragraph{Error handling.} LLM calls are retried up to 5 times with exponential backoff on transient errors, excluding malformed-request errors, authentication errors, and context/output-length errors (which are handled separately rather than retried). If the context window is exceeded, the harness falls back to summarizing the trajectory and unwinding older messages to free tokens before retrying; if the model's output is truncated, the harness informs the model and retries once. An Anthropic-specific prompt-caching helper adds ephemeral \texttt{cache\_control} hints to the most recent messages, but is a no-op for these experiments since it only activates when the model name contains ``anthropic''/``claude,'' and the shared \texttt{AGENT\_MODEL} is \texttt{openai/gpt-5.5}.

\paragraph{System prompt.} The system prompt frames the task as command-line problem solving in a Linux environment, explicitly states the agent has no human intervention and no way to perceive multimedia directly (motivating the \texttt{image\_read} tool), and instructs the agent, before calling \texttt{task\_complete}, to identify the minimal set of files that must change and verify that no other state was altered.

\subsection{GEPA}
GEPA optimizes only the system prompt; all other harness components (tools, error handling, control code) are unchanged from the baseline agent. The prompt grows from 5 lines (initial) to 103 lines (Phase 1) to 195 lines (Phase 2).

\paragraph{Phase 1 changes.} The Phase 1 prompt adds a generic "recommended workflow" checklist, plus a set of "important lessons from prior attempts" organized as three named per-task lessons, each pairing a Terminal-Bench task ID with exact file paths, literal error strings, and step-by-step fixes:
\begin{itemize}
    \item \texttt{configure-git-webserver}: git/web-server configuration, including the exact document-root path and HTTP status expected by the verifier (Appendix~C).
    \item \texttt{torch-tensor-parallelism}: the expected weight-partitioning behavior and forward-pass contract for a column/row-parallel PyTorch linear layer.
    \item \texttt{torch-pipeline-parallelism}: expected microbatching, send/recv ordering, and backward-pass behavior for a PyTorch pipeline-parallel implementation.
\end{itemize}

\paragraph{Phase 2 changes.} The Phase 2 prompt keeps all three Phase 1 lessons and adds two more, in the same style:
\begin{itemize}
    \item \texttt{mcmc-sampling-stan}: literal RStan sampling output patterns (e.g., \texttt{"SAMPLING FOR MODEL"}, \texttt{"Chain"}) that the verifier checks for in stdout.
    \item \texttt{make-doom-for-mips}: the exact expected stdout initialization string and output-image dimensions for a cross-compiled DOOM build.
\end{itemize}

\paragraph{Excluded variant.} The code-mutation variant (GEPA-code) is not included here, since it failed to produce a valid candidate during Phase 1.

\subsection{Meta Harness}
Meta Harness edits harness code directly; the prompt template is unchanged across phases.

\paragraph{Phase 1 candidate (\texttt{io\_boundary\_hardening.py}, 1315 lines, up from 1248 in the initial harness).} Replaces substring matching for command-completion markers with a regex (\texttt{\_MARKER\_RE}) plus explicit handling of marker-suffix collisions, and adds defensive type coercion for malformed tool-call arguments (e.g., non-numeric durations, non-string keystrokes) with logged warnings in place of silent failure.

\paragraph{Phase 2 candidate (\texttt{output\_noise\_compaction.py}, 1474 lines).} Adds a pass that categorizes long, repetitive terminal output (package-manager install logs, download/build progress bars, test-runner boilerplate) by regex and collapses each repeated run into a short summary marker with a few representative examples, while preserving a raw tail of the original output rather than discarding it outright.

Neither candidate contains task-specific literals or task IDs.

\subsection{RELAI-VCL}
RELAI-VCL edits the full agent package (prompt template, tool definitions, and control code) in both phases.

\paragraph{Phase 1 candidate.} Adds a keyword-based classifier that distinguishes code-editing tasks from operational/service tasks; a completion gate that runs verifier-discovery and runtime-contract-check commands before accepting \texttt{task\_complete}; session-loss recovery handling (capturing and returning the last visible terminal state if the tmux session dies mid-command); and defensive handling of malformed tool-call arguments.

\paragraph{Phase 2 candidate.} Adds fatal-runtime-error detection (tracebacks, missing packages, nonzero exit codes) that blocks completion until the underlying error is resolved; a check for absolute workspace-path references in generated code that would break under chroot/jail-isolated verifiers; and an expanded search over verifier file locations, including absolute \texttt{/tests/...} paths.

The prompt template itself is fully generic in both phases, with no task IDs or literal expected outputs.

\section{Method-Specific Diffs and Failure Cases}
Appendix~B summarizes what each method's accepted candidates actually contain. This section gives one short excerpt per method to make those descriptions concrete, followed by the two specific outcomes referenced in the main text. Each excerpt below is a small, representative fragment of a much larger change, not the entire diff; the full prompts and code for every phase of every method are released in the accompanying artifact repository (see title-page footnote).

\paragraph{GEPA.} The excerpt below reproduces the start of one full named lesson from the Phase 1 prompt, including the header that names the task ID (\texttt{configure-git-webserver}) directly, followed by its first bullet. The Phase 1 prompt contains three such named lessons in total (Appendix~B.2), of which this is one:

\begin{instructbox}[Excerpt: GEPA Phase 1 prompt (partial)]
Important lessons from prior attempts:\\
\\
A. Web server / Git configuration tasks, especially `configure-git-webserver`\\
- Always inspect `/tests/verify.sh` before configuring the system. The verifier may check exact URLs, file paths, package availability, service status, and expected file contents.\\
- A previous failure returned HTTP 404 when testing the web server. Avoid this by ensuring the required file, commonly `hello.html`, exists in the actual document root served by the active web server, typically `/var/www/html/hello.html` for Apache/nginx on Ubuntu.\\
{[}...{]}
\end{instructbox}

\paragraph{Meta Harness.} The Phase 1 candidate replaces simple substring matching for command-completion markers with a regex plus explicit suffix-collision handling, a generic robustness fix with no task-specific content. The excerpt below is one fragment of this change:

\begin{lstlisting}[language=Python]
_MARKER_RE = re.compile(r"__CMDEND__\d+__")
...
marker_suffixes = {
    marker[i:]
    for marker in markers
    for i in range(len(marker))
    if marker[i:].endswith("__") and any(ch.isdigit() for ch in marker[i:])
}
lines = [
    line
    for line in lines
    if not _MARKER_RE.search(line) and line.strip() not in marker_suffixes
]
\end{lstlisting}

\paragraph{RELAI-VCL.} The excerpt below is the complete text of one workflow note added in Phase 1 (one part of the larger Phase 1 change described in Appendix~B.4), used to route operational/infrastructure tasks toward verifier-shaped validation; it is representative of RELAI-VCL's changes in containing no task IDs or literal expected outputs:

\begin{lstlisting}[language=Python]
OPERATIONAL_TASK_WORKFLOW_NOTE = """
For operational/infrastructure tasks, use a verifier-shaped delivery loop:
1. Infer the externally visible deliverable, not just the process liveness condition.
2. Inspect nearby task files, tests, verifier scripts, or obvious validation assets when present so you know what a checker will request over the network, filesystem, or CLI.
3. Treat HTTP 200 on a root page, an open port, a running service, or a clean daemon status as intermediate evidence unless the task explicitly says that is the final deliverable.
4. Before finishing, perform one end-to-end audit that exercises the published artifact or workflow result the verifier is likely to check.
5. If the task depends on a user workflow step such as clone/push/fetch/request, simulate that workflow or validate the exact resulting externally served artifact before stopping.
"""
\end{lstlisting}

\begin{itemize}
    \item \textbf{GEPA-code's invalid-candidate outcome (Section~4.2).} The code-mutation variant of GEPA produced no valid candidate during Phase 1 and is excluded from all Phase 2 comparisons; no corresponding artifact exists for it.
    \item \textbf{Meta Harness's Phase 2 stall (Section~6.3).} Every candidate generated during Meta Harness's second optimization round scored worse than its Phase-1-optimized agent, so its Phase 2 pass rate on the combined task set fell below its transfer pass rate. The one Phase 2 harness file released here (\texttt{output\_noise\_compaction.py}) is the candidate that was ultimately kept; the released artifacts do not include the rejected candidates from that round.
\end{itemize}

\end{document}